\documentclass[10pt,twocolumn,letterpaper]{article}

\usepackage[pagenumbers]{cvpr} 

\usepackage{multirow}
\newcommand{\projecttitle}{MultiGO~}
\newcommand{\myparagraph}[1]{\smallskip \noindent{\bf {#1}.}}

\definecolor{cvprblue}{rgb}{0.21,0.49,0.74}
\usepackage[pagebackref,breaklinks,colorlinks,allcolors=cvprblue]{hyperref}

\def\paperID{10038} 
\def\confName{CVPR}
\def\confYear{2025}

\title{MultiGO: Towards Multi-level Geometry Learning for Monocular \\3D Textured Human Reconstruction}

\author{Gangjian Zhang\textsuperscript{1}, Nanjie Yao\textsuperscript{1}, Shunsi Zhang\textsuperscript{2}, Hanfeng Zhao\textsuperscript{2}, \\ 
Guoliang Pang\textsuperscript{2}, Jian Shu\textsuperscript{1}, Hao Wang\textsuperscript{1} \\
\textsuperscript{1}\textit{The Hong Kong University of Science and Technology (Guangzhou)}, China \\
\textsuperscript{2}\textit{Guangzhou Quwan Network Technology}, China \\
Email: gzhang292@connect.hkust-gz.edu.cn, nanjiey@uci.edu, haowang@hkust-gz.edu.cn \\
\url{https://multigohuman.github.io/}
}

\begin{document}



\twocolumn[{%
\maketitle
\begin{center}
    \vspace{-0.1in}
    \centering
    \captionsetup{type=figure}
    \includegraphics[width=0.95\textwidth]{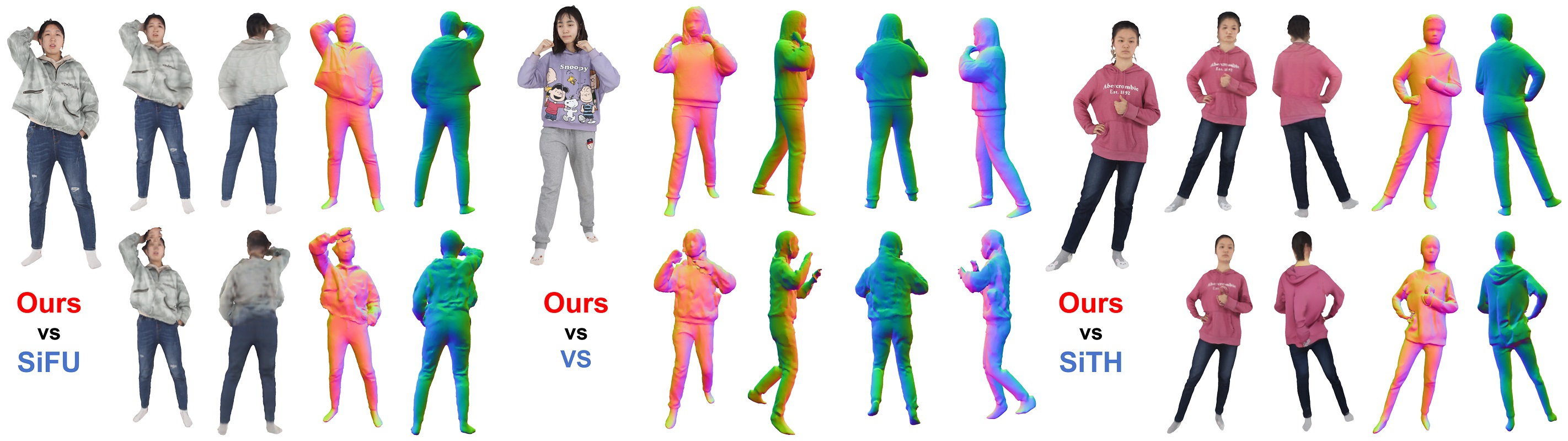}
    \vspace{-6pt}
    \vspace{-0.1cm}
    \captionof{figure}{\textbf{Comparisons with the SOTA Methods in Monocular 3D Textured Human Reconstruction.} Existing SOTA methods struggle with recovering correct human poses and intricate geometry details. SiFU~\cite{Zhang_2024_sifu} is unable to reconstruct correct human postures, such as incorrect left-hand positions. VS~\cite{VS_CVPR2024} performs poorly in fine-grained areas such as unclear finger movement and cloth wrinkles. SiTH~\cite{ho2024sith} produces geometry and texture errors that occur from the generative model, such as the third arm on the back. }
    \label{fig:enter-label}
\end{center}%

}]

\begin{abstract}

This paper investigates the research task of reconstructing the 3D clothed human body from a monocular image. Due to the inherent ambiguity of single-view input, existing approaches leverage pre-trained SMPL(-X) estimation models or generative models to provide auxiliary information for human reconstruction. However, these methods capture only the general human body geometry and overlook specific geometric details, leading to inaccurate skeleton reconstruction, incorrect joint positions, and unclear cloth wrinkles. In response to these issues, we propose a multi-level geometry learning framework. Technically, we design three key components: skeleton-level enhancement, joint-level augmentation, and wrinkle-level refinement modules. Specifically, we effectively integrate the projected 3D Fourier features into a Gaussian reconstruction model, introduce perturbations to improve joint depth estimation during training, and refine the human coarse wrinkles by resembling the de-noising process of diffusion model. Extensive quantitative and qualitative experiments on two out-of-distribution test sets show the superior performance of our approach compared to state-of-the-art (SOTA) methods. 


\end{abstract}    
\section{Introduction}
\label{sec:intro}

With the increasing popularity of virtual worlds, there is a growing demand for the realistic digital human creation. To achieve it efficiently, monocular 3D human reconstruction has been an important task.
However, since single-view images cannot provide sufficient information for reconstruction, there exists substantial ambiguity in simulating the geometry and texture of the occluded human body parts.

To address this issue, existing methods have explored the introduction of pre-conditioned SMPL-X techniques to provide a 3D body mesh as geometric prior information. These methods employ various options for human body geometry priors, such as SMPL-X normal maps~\cite{xiu2022icon,xiu2023econ, Zhang_2024_sifu, VS_CVPR2024}, low-frequency and high-frequency signals of the SMPL-X body~\cite{hilo}, triplane representaion~\cite{zhang2024global_gta}, and the outputs of generative models~\cite{saito2019pifu,ho2024sith}, in an attempt to achieve more accurate human body reconstruction. 

However, these traditional approaches often focus on modeling the general geometry of the human body, overlooking the multi-level structures that include the skeleton, joints, and finer details such as wrinkles around the fingers and face. This oversimplified modeling approach results in inaccurate skeletal reconstructions, incorrect joint positions, and unclear representation of clothing wrinkles.


Moreover, these approaches often utilize 3D representation methods like occupancy grids~\cite{saito2019pifu}, SDF~\cite{sdf}, and NeRF~\cite{mildenhall2021nerf}. While these representations are capable of accurately modeling human body geometry, they are often burdened by high computational costs and low efficiency. 
Recently, 
Szymanowicz et al.~\cite{szymanowicz2023splatter} perform monocular reconstruction by mapping a single object image into 3D Gaussian points using a neural network in a feed-forward process, which enables effective and efficient 3D object reconstruction. This work motivates us to explore Gaussian models for the monocular human reconstruction task.


To this end, this paper proposes a novel multi-level geometry learning framework, MultiGO, based on an existing object Gaussian reconstruction model~\cite{tang2024lgm}. We aim to enhance the geometry of human reconstructions across various levels of granularity, including skeletons, joints, and wrinkles, thereby largely improving 3D human reconstruction quality. The proposed MultiGO contains three key components, i.e., the Skeleton-Level Enhancement (SLE) module, Joint-Level Augmentation (JLA) strategy, and  Wrinkle-Level Refinement (WLR) module, which deal with the specificity of three different levels of the human geometry respectively.

Technically, (1) the SLE module is designed to enhance the accuracy of capturing human overall posture by effectively bridging the 3D SMPL-X prior with 2D human imagery. By projecting 3D Fourier features into the same 2D space as the input image, the SLE module allows the model to fully utilize established geometric priors related to human shapes and configurations. (2) The JLA strategy tackles the significant challenge in estimating the depth of human joints in 3D space. It recognizes that, during inference, inaccuracies in the estimated SMPL-X depth prior can lead to substantial errors. To address this, the JLA strategy introduces controlled perturbations to the ground truth joint positions during training. (3) The WLR module refines geometric details, such as wrinkles on the human body, by using high-quality textures generated from Gaussian representations as a conditioning. It effectively resembles this refinement process to the final stage of diffusion de-noising. 

Experiments on two out-of-distribution (OOD) test sets, CustomHuman and THuman3.0 validate that our proposed multi-level geometry learning framework achieves SOTA performance. Our contributions are as follows:
\begin{itemize}
    
    \item A Skeleton-Level Enhancement module, which captures human overall posture by integrating the projected 3D Fourier features with 2D images.

    \item A Joint-Level Augmentation strategy, which applies perturbations to joint positions during training to improve the model's resilience to depth prior inaccuracy in inference.

    \item A Wrinkle-Level Refinement module, which refines the coarse geometric wrinkles using reconstructed Gaussian texture in a de-noising manner of the diffusion model.

\end{itemize}

\section{Related Work}
\label{sec:rw}

\subsection{Monocular Human Reconstruction}

Monocular human reconstruction has become a popular research focus for human digitalization. The pioneering method, PIFu~\cite{saito2019pifu}, introduced a pixel-aligned implicit function for shape and texture generation. ICON~\cite{xiu2022icon} improved on this by using skinned body models~\cite{loper2023smpl} as human body priors, while ECON~\cite{xiu2023econ} combined implicit representations with explicit body regularization. GTA~\cite{zhang2024global_gta} used a 3D-decoupling transformer and hybrid prior fusion for comprehensive 3D geometry and texture reconstruction. VS~\cite{VS_CVPR2024} introduced a "stretch-refine" strategy for managing large deformations in loose clothing. HiLo~\cite{hilo} enhanced geometry details by extracting high and low-frequency information from a parametric model, improving noise robustness as a result. SiTH~\cite{ho2024sith} tackled occlusion by leveraging a 2D prior from the SD model. Lastly, SiFU~\cite{Zhang_2024_sifu} utilized a cross-attention mechanism in transformers to optimize geometry, employing 2D diffusion models for texture prediction.
 

\subsection{Human Gaussian Model}

Recent advancements in 3D Gaussian splatting~\cite{3DGaussian} have promoted 3D human creation. Traditional representations like SDF and NeRF often struggle with balancing efficiency and rendering quality. Techniques such as HuGS~\cite{HuGS}, D3GA~\cite{D3GA}, and 3DGS-Avatar~\cite{3DGS-Avatar} utilize rich spatial and temporal data for modeling in multi-view videos, monocular video sequences, and sparse-view inputs. Animatable Gaussians enhance garment dynamics through pose projection mechanisms and 2D CNNs. Notably, Gauhuamn~\cite{hu2024gauhuman} and HUGS~\cite{HuGS} optimize human Gaussians from monocular videos, while HiFi4G~\cite{Hifi4g} employs a dual-graph mechanism for spatial-temporal consistency, and ASH~\cite{Ash} uses mesh UV parameterization for real-time rendering. GPS-Gaussian~\cite{GPS-Gaussian} proposes a generalizable multi-view human Gaussian model with high-quality rendering.

\subsection{Large 3D Object Reconstruction Model }  

\begin{figure*}
    \centering
    \includegraphics[width=1.0\linewidth]{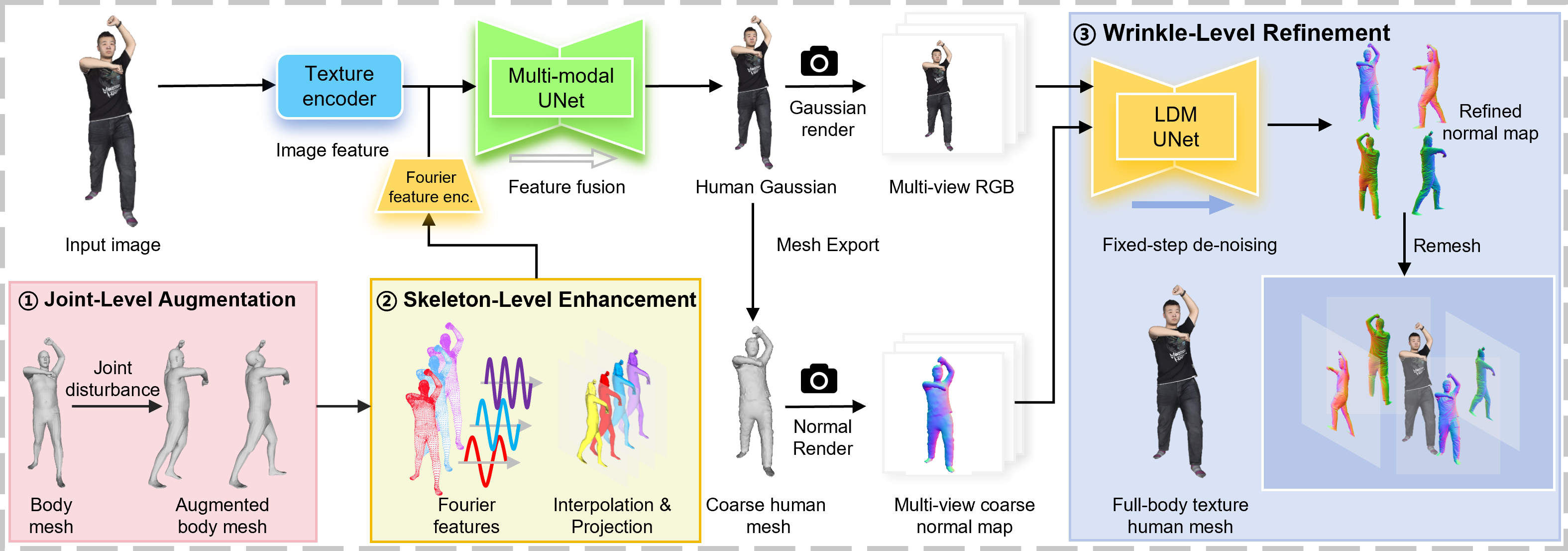}
    \caption{\textbf{Method Overview.} Our method, MultiGO, addresses monocular textured 3D human reconstruction by introducing a multi-level geometry learning framework that significantly enhances reconstruction quality. To accurately capture the human body's posture, we propose the SLE module, which projects 3D Fourier features into the 2D space of the input image, allowing the Gaussian reconstruction model to fully utilize prior human shape knowledge. For improved depth estimation of human joints, the JLA strategy applies controlled perturbations during training, increasing the model's robustness to depth inaccuracies during inference. To refine geometric details like body wrinkles, the WLR module resembles the final de-noising steps in diffusion theory, treating coarse meshes as Gaussian noise and using the high-quality texture of reconstructed Gaussian as conditions to refine wrinkles.} 
    \label{fig: pipeline}
    \vspace{-0.2cm}
\end{figure*}

Recent advancements in 3D object reconstruction leverage large models with convolutional or transformer backbones for efficient 2D-to-3D conversions. LRM~\cite{hong2023lrm} has notably enhanced model capacity and data volume, enabling direct 3D reconstruction from a single image. Further improvements, such as instant3d and others~\cite{li2023instant3d, xu2023dmv3d, wang2023pf}, utilize multi-view diffusion models for better results. Key datasets like Objaverse~\cite{deitke2023objaverse} support training these models, with significant focus on 3D Gaussian representation~\cite{tang2024lgm, xu2024grm, zhang2024gslrm, xu2024agg} and 3D triplane representation~\cite{wang2024crm}. These methods often require extensive 3D data for pre-training diffusion models~\cite{wang2023imagedream, long2023wonder3d, shi2023zero123++}, which are crucial for generating novel multi-view inputs. 
\section{Methodology}
\label{sec:method}


\subsection{Preliminaries}
\label{prel}
\myparagraph{Gaussian Splatting} Gaussian splatting~\cite{3DGaussian}, emerging as a popular 3D representation, utilize a collection of 3D Gaussians, denoted by $\Theta$, to model 3D data. Each 3D Gaussian is characterized by a parameter set: $\theta_i = {\mathbf{x}_i, \mathbf{s}_i, \mathbf{q}_i, \alpha_i, \mathbf{c}_i} \in \mathbb{R}^{14}$. Here, $\mathbf{x} \in \mathbb{R}^3$ represents the geometry center, $\mathbf{s} \in \mathbb{R}^3$ the scaling factor, $\mathbf{q} \in \mathbb{R}^4$ the rotation quaternion, $\alpha \in \mathbb{R}$ the opacity value, and $\mathbf{c} \in \mathbb{R}^3$ the color feature. In our work, we set the dimension of $\mathbf{c}$ to 3 for rendering purposes, allowing spherical harmonics to model view-dependent effects.


\myparagraph{SMPL Series Model} The Skinned Multi-Person Linear (SMPL) model~\cite{loper2023smpl} is a parametric model to represent human body. It utilizes a set of body parameters $\mathcal{B} \in \mathbb{R}^{d}$ to define a given human body mesh $\mathcal{M}$:
\begin{equation}
    \mathcal{M}(\mathcal{B}): \mathcal{B} \Rightarrow \mathbb{R}^{3\times6890}.
\end{equation}
Each parameter in $\mathcal{B}$ controls the position or orientation of body parts, etc. Many extensions such as SMPL-X~\cite{loper2023smpl}, and SMPL-H~\cite{romero2022embodied_smplh} add more parameters for facial expressions, finger movement, and other fine-grained poses. By default, we use SMPL-X as our human body model in this paper.

\myparagraph{Latent Diffusion Model} 
Diffusion model~\cite{ho2020denoisingdiffusionprobabilisticmodels} proposes to generate images through a degradation process. The latent diffusion model (LDM)~\cite{rombach2022high} incorporates a pre-trained variational auto-encoder (VAE)~\cite{vae}, including the encoder and decoder. Starting with a sample $z_0$ from the VAE latent distribution $z$, the forward process produces a sequence of noised data $\{z_t \mid t \in (0, T)\}$, where $z_t = z_0 + \epsilon$. Here $\epsilon$ is randomly sampled noise from a Gaussian noise $\mathcal{N}(0,1)$. Conversely, the reverse process uses an iterative de-noising way to recover $z_{t-1}$ from $z_t$ by predicting the noise $\epsilon$.

\subsection{The Proposed Method}
\subsubsection{Overview}
Our method, illustrated in Figure~\ref{fig: pipeline}, processes the human front-view 
RGB image and a 3D body mesh generated by the SMPL-X model. A texture encoder extracts 2D image features from the RGB input, while the 3D mesh is transformed into geometric features via our SLE module. These features are then integrated using a multimodal UNet, resulting in 3D human Gaussian points. The SLE module and fusion process are detailed in Section~\ref{SLE}. To enhance depth estimation accuracy during inference, we introduce a JLA strategy, explained in Section~\ref{JLA}. Additionally, to improve mesh detail post-Gaussian export, we propose the WLR module, described in Section~\ref{sec:HMRM}.



\begin{figure}[t!]
    \centering
    \includegraphics[width=1.0\linewidth]{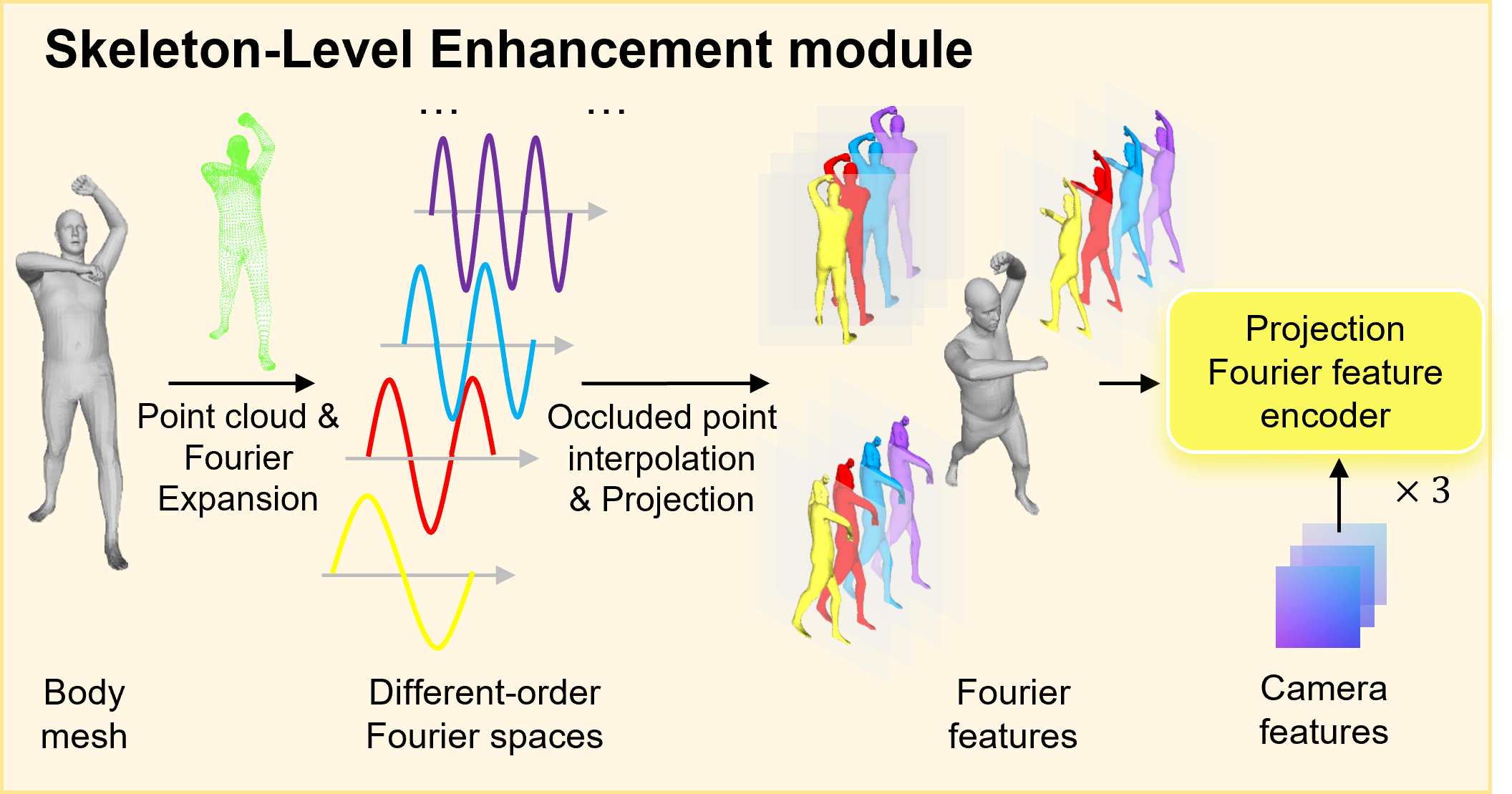}
    \caption{\textbf{Skeleton-Level Enhancement Module.} To enhance the human geometry at the skeleton level, we achieve better fusion of the heterogeneous modalities of the 3D SMPL-X body prior and 2D images. We propose interpolating the Fourier features of 3D occluded points and mapping them from three different angles into the same 2D space as the image features. }
    \label{fig: SLE}
    \vspace{-0.3cm}
\end{figure}

\subsubsection{Skeleton-Level Enhancement Module} \label{SLE}
In the previous analysis, the monocular setting of this task means a frontal human RGB image alone cannot provide sufficient geometric information. To overcome this, we propose a novel Skeleton-Level Enhancement (SLE) module that effectively incorporates the 3D geometric prior knowledge of the SMPL-X body mesh. Recognizing that image features and 3D SMPL-X features originate from two different modalities with a huge semantic gap, our approach avoids the rigid fusion of these features. Instead, the SLE module projects 3D Fourier features into the same 2D space as the input image, facilitating better interaction and fusion between the heterogeneous features. This module enables the model to effectively learn the geometric skeleton of the human, even with a limited amount of human data. 




 Concretely, as shown in Figure~\ref{fig: SLE}, inspired by some works~\cite{li2024craftsman,3DShape2VecSet} the proposed SLE module first considers all vertices of the input body mesh as points of the point cloud. The body point cloud can be represented as $\mathcal{P}_{0} \in \mathbb{R}^{3\times10475}$. Then, the 3D Fourier expansion operation is used to enhance the expression of these points. Specifically, we extract $q$-order Fourier series for each point $p$ in $\mathcal{P}_{0}$ as follows:
\begin{equation} \label{mand1-m}
    \mathcal{F}(p)  = \left\{p\right\} \cup \left\{\cos(2^{n}p), \sin(2^{n}p)|n \in\left\{1,...,q\right\}\right\}.
\end{equation}
Through the above operation, we have expanded the 3D space where the original SMPL-X points are located into the different Fourier spaces $\left\{\mathcal{S}_{n} | n \in\left\{0,...,2q\right\} \right\}$. The point clouds in these spaces are denoted as $\left\{\tilde{\mathcal{P}_{n}} | n \in\left\{0,...,2q\right\} \right\}$. Meanwhile, to make the points in these spaces denser, we have interpolated and expanded them. Specifically, we select positions on the surface of a triangular surface and average the weights of three vertices belonging to the same triangular surface. After this, denser point clouds $\tilde{\mathcal{P}_{n}} \in \mathbb{R}^{3\times m}$ with different-order Fourier are obtained, where $m$ is the point number.

To facilitate the fusion of geometric and texture features, we perform 2D projection on the occluded points in different Fourier spaces from three camera angles. By doing so, we can obtain a stack of Fourier features from different spaces as shown in Figure~\ref{fig: SLE}, which can be concatenated into $\tilde{\mathcal{F}_{1}} \in \mathbb{R}^{3(2p+1) \times H \times W}$, where $H$ and $W$ are the projection resolution. Similarly, from the perspectives of the other two cameras, we can obtain $\tilde{\mathcal{F}_{2}}$ and $\tilde{\mathcal{F}_{3}}$. Subsequently, all of them, along with their camera feature, are fed into a Fourier feature encoder to obtain geometric features, $\mathcal{F}_{1}^{\prime}, \mathcal{F}_{2}^{\prime} \mathcal{F}_{3}^{\prime} \in \mathbb{R}^{o \times h \times w}$. We design convolution layers to compose the Fourier feature encoder, making it output the same dimension as the image latent $\mathcal{I}_{0}^{\prime} \in \mathbb{R}^{o \times h \times w}$ obtained after texture encoding the RGB image $\mathcal{I}_{0} \in \mathbb{R}^{3 \times H \times W}$.

In our approach, features from two modalities and four perspectives are combined using a multimodal fusion UNet~\cite{tang2024lgm}, enhanced with residual and self-attention layers. The self-attention layers serve to facilitate interaction between different perspectives and integrate features across modalities. These fused features are then decoded to predict the final textured Gaussian parameters, $\Theta$, forming the ultimate feature map. Each pixel is modeled as a 3D Gaussian $\theta_i \in \mathbb{R}^{14}$. We utilize the differentiable renderer~\cite{xu2024grm} to render these Gaussians. RGB and alpha images from eight views, including one input and seven novel views, are rendered. The loss is computed by comparing these rendered images to GT scans, using MSE and VGG-based LPIPS loss~\cite{zhang2018unreasonable} for RGB images, and MSE loss for alpha images.

\begin{figure}
    \centering

    \includegraphics[width=1\linewidth]{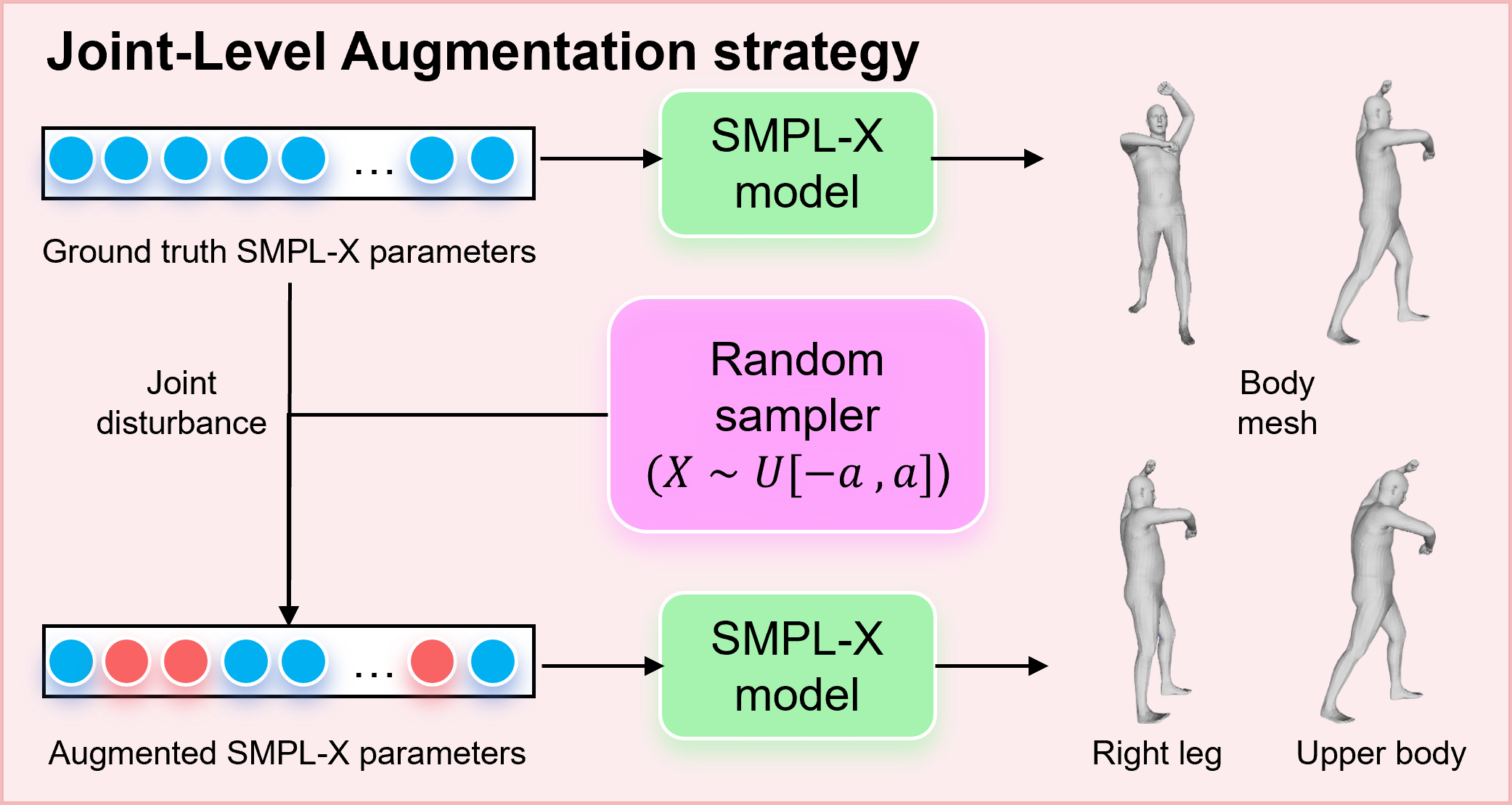}
    \caption{\textbf{Joint-Level Augmentation Strategy.} To enhance human geometry at the joint level, we augment the samples of input human SMPL-X body mesh during training. We propose to randomly perturb ground truth SMPL-X parameters associated with specific joints to increase the model robustness in inference.}
    \label{fig: JLA}

\end{figure}

\subsubsection{Joint-Level Augmentation Strategy} \label{JLA}

The SLE module aids in fitting the model to overall human pose inputs effectively. However, unlike the precise SMPL-X body parameters from human scans used during training, those used during inference are estimated from a single view, leading to potential inaccuracies in reflecting true human geometry. This results in depth discrepancies in the fitted body mesh, for example, the difference in the front and back position of the feet is obvious when viewed from the side, but not obvious when viewed from the front. To mitigate these inaccuracies in SMPL-X depth estimation during inference, we propose a Joint-Level Augmentation (JLA) Strategy, which involves altering some body parameters in the training data, as illustrated in Figure~\ref{fig: JLA}.

To simplify, we represent the scan-fitted GT human body parameters as a vector, $\mathcal{B} = (\beta^{0}, \beta^{1}, \ldots, \beta^{d}) \in \mathbb{R}^{d}$, where $d$ is the total number of body parameters. Each element in this vector influences posture changes, for example, $\beta^{0}$ controls the forward and backward rotation of the left leg, and $\beta^{35}$ manages the left and right rotation of the head. As illustrated in Figure~\ref{fig: JLA}, these parameters are input into a pre-trained SMPL-X model to generate a standard body mesh.

During training, instead of using these standard body meshes directly, we propose creating augmented samples to simulate inference scenarios. Specifically, we focus on parameters affecting depth information, like $\beta^{0}$, which are more likely to produce errors in single-view SMPL-X estimation during inference. We construct a body depth-related mask vector $M = (\mu^{1}, \mu^{2}, \ldots, \mu^{d}) \in \mathbb{R}^{d}$, where each element $\mu^{j} \in {0, 1}$ for $j \in {1, 2, \ldots, d}$. For elements where $\mu^{j} = 1$, we generate a sample from a uniform distribution $X \sim U[-\alpha, \alpha]$ to produce an offset $x^{j}$, with $\alpha$ as a hyperparameter controlling the offset degree. By adding these offsets to the GT SMPL-X parameters, we obtain the augmented SMPL-X parameters $\tilde{\beta} = (\tilde{\beta}^{0}, \tilde{\beta}^{1}, \ldots, \tilde{\beta}^{d}) \in \mathbb{R}^{d}$. Each $\tilde{\beta}^{j}$ is calculated as follows:
\begin{equation} \label{mand1-m}
    \tilde{\beta}^{j} = \beta^{j} + \mu^{j} * x^{j}
\end{equation}
The augmented parameter vector is input into the pre-trained SMPL-X model to produce a body mesh with minor geometric joint changes from the standard mesh. This process helps the model develop correction abilities, ensuring accurate predictions even when discrepancies exist between the input SMPL-X mesh and the target human geometry. This correction ability is derived from extracting and understanding image modality.



\subsubsection{Wrinkle-Level Refinement Module} \label{sec:HMRM}
\begin{figure}
    \centering
        \vspace{-0.1cm}

    \includegraphics[width=1\linewidth]{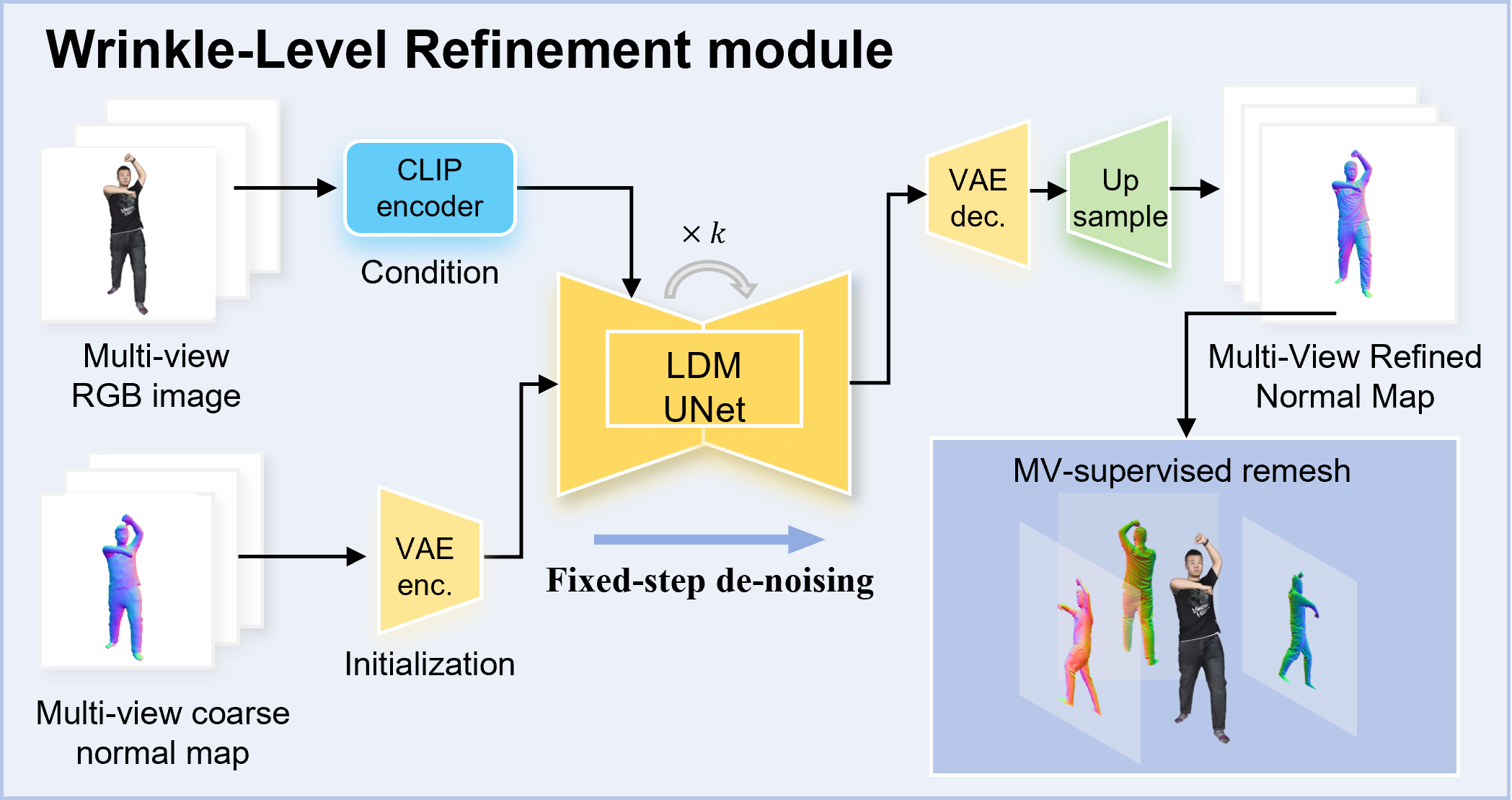}
    \caption{\textbf{Wrinkle-Level Refinement Module.} To improve the human geometry at the wrinkle level, we equate the refinement process with the last few steps of the de-noising process in the diffusion model and use a fixed number of ``de-noising" steps to achieve the refined mesh predicting from initialized coarse mesh. }
    \label{fig:WLR}
    \vspace{-0.3cm}
\end{figure}

While the SLE module and JLA strategy allow for the creation of high-quality 3D Gaussian point clouds of the human body, converting these into meshes using the Gaussian-to-mesh-export technique~\cite{tang2024lgm} often fails to capture fine details like facial expressions and clothing. To enhance the geometric details of these meshes, we introduce the Wrinkle-Level Refinement (WLR) module. This module utilizes the detailed texture of the reconstructed Gaussian as the condition to refine mesh wrinkles. Additionally, given our access to a coarse mesh and a near-ground-truth (GT) mesh, our approach innovatively resembles the refinement process with the final de-noising stage in diffusion theory. We treat the coarse mesh as if it has been subjected to $k$ iterations of Gaussian noise, then apply diffusion and de-noising pipeline to achieve enhanced results.


The full WLR is in Figure~\ref{fig:WLR}. Starting with a reconstructed human Gaussian $\mathcal{G}$ and a coarse human mesh $\mathcal{M}^{c}$, we use differentiable renderers $\textit{R}_{\mathcal{G}}$ and $\textit{R}_{\mathcal{M}^{c}}$ to generate the Gaussian-rendered color image $I_{c}=\{I_{c}^1,..., I{c}^N\}$ and mesh-rendered coarse normal map $I_{n}=\{I_{n}^1,..., I_{n}^N\}$from a common set of camera views \textbf{k}. Then, the pre-trained CLIP~\cite{ramesh2022hierarchical} image encoder and VAE encoder $\mathcal{E}$ are employed to encode the color image and normal map, respectively.

\myparagraph{Fine-tuning and inference}
Training a data-driven diffusion model from scratch is challenging, especially with limited human data. To address this, we fine-tune a pre-trained latent diffusion model UNet~\cite{unique3d}, denoted as $\epsilon_{\theta}$. During fine-tuning, we define the latent of the ground truth (GT) normal map as $z_0 = \mathcal{E}(I_n')$ and the latent of the coarse normal map as $z_k = \mathcal{E}(I_n)$, which can be taken as the result of adding noise $k$ times to $z_0$ from the traditional perspective of diffusion. The model $\epsilon_{\theta}$ is to predict the distribution discrepancy $\epsilon$ between $z_0$ and $z_k$. And the time step $t=k$. The objective function for our training is as follows:



\begin{equation}
    \mathop{\min}_{\theta} \mathbb{E}_{\epsilon} \Vert\epsilon - \epsilon_{\theta}(z_k, k, I_{c} )\Vert_{2}^{2}
\end{equation}
Then during inference time, we can obtain the refined normal map $\hat{I_{n}}$ by:
\begin{equation}
    \label{eq:func}
    \hat{I_{n}} = \mathcal{D}(\hat{z_0}) = \mathcal{D}(\mathcal{F}(z_0, I_{c}))
\end{equation}
where $\mathcal{D}$ is the VAE decoder, $\hat{z_0}$ is the refined latent, and the $\mathcal{F}$ represents the $k$-step de-noising process of our model. Additionally, we applied a fine-tuned super-resolution model~\cite{realesrgan} for upsampling, which allows a refined human normal map with intricate geometric details. 

\begin{table*}[t!]
    \centering
    \renewcommand{\arraystretch}{0.85}
    \scalebox{0.8}{
    \begin{tabular*}{\textwidth}{@{\extracolsep{\fill}}lc|ccc|ccc}
    \toprule
         \multirow{2}{*}{Methods}  & \multirow{2}{*}{Publication}     & \multicolumn{3}{c}{CustomHuman~\cite{ho2023customhuman}} & \multicolumn{3}{c}{THuman3.0~\cite{thuman3.0}}  \\
                                   &                               & \begin{tabular}{c} CD: P-to-S /\\ S-to-P $(\mathrm{cm}) \downarrow$ \end{tabular} & NC $\uparrow$ & f-score $\uparrow$ &
                                                                    \begin{tabular}{c} CD: P-to-S /\\ S-to-P $(\mathrm{cm}) \downarrow$ \end{tabular} & NC $\uparrow$ & f-score $\uparrow$  \\
    \midrule 
    PIFu~\cite{saito2019pifu}        &  ICCV 2019  & $2.965/3.108$ & 0.765 & 25.708 & $2.176/2.452$  & 0.773 & 34.194 \\
    GTA~\cite{zhang2024global_gta}   & NeurIPS 2023& $2.404/2.726$ & 0.790 & 29.907 & $2.416/2.652$   & 0.768 & 29.257 \\    
    ICON~\cite{xiu2022icon}          & CVPR 2022   & $2.437/2.811$ & 0.783 & 29.044 & $2.471/2.780$   & 0.756 & 27.438 \\
    ECON~\cite{xiu2023econ}          & CVPR 2023   & $2.192/2.342$ & 0.806 & 33.287 & $2.200/2.269$   & 0.781 & 33.220 \\   
    VS~\cite{VS_CVPR2024}            & CVPR 2024   & $2.508/2.986$ & 0.779 & 26.790 & $2.523/2.943$   & 0.758 & 26.340  \\    
    HiLo~\cite{hilo}                 & CVPR 2024   & $2.280/2.739$ & 0.794 & 30.275 & $2.385/2.862$   & 0.765 & 28.119 \\
    SiFU~\cite{Zhang_2024_sifu}      & CVPR 2024   & $2.464/2.782$ & 0.790 & 28.574 & $2.480/2.822$   & 0.762 & 27.929 \\
    SiTH~\cite{ho2024sith}           & CVPR 2024   & $1.800/2.188$ & 0.816 &36.148 & $1.763/2.002$ &0.787  &36.230 \\
    \projecttitle                    &     -       & $\textbf{1.620}/\textbf{1.782}$ & \textbf{0.850} & \textbf{42.425} & $\textbf{1.408}/\textbf{1.633}$ & \textbf{0.834} & \textbf{46.091} \\
    \bottomrule
    \end{tabular*}
    }
    \vspace{-0.1cm}

    \caption{\textbf{Comparison of Human Geometry with SOTA methods.}  The \textbf{best} results are highlighted with bold. Arrow $\uparrow$/$\downarrow$ means higher/lower is better. Note that all the experiments are tested with the estimated SMPL/SMPL-X model instead of the GT one.\label{main_exp_3d}}
\end{table*}

\begin{table*}[t!]
    \centering
    \renewcommand{\arraystretch}{1}
        \scalebox{0.8}{
    \begin{tabular*}{\textwidth}{@{\extracolsep{\fill}}l|ccc|ccc}
    \toprule
         \multirow{2}{*}{Methods}  & \multicolumn{3}{c}{CustomHuman} & \multicolumn{3}{c}{THuman3.0}  \\
                                   & LPIPS: F/B $\downarrow$ & SSIM: F/B $\uparrow$ & PSNR: F/B $\uparrow$ &
                                     LPIPS: F/B $\downarrow$ & SSIM: F/B $\uparrow$ & PSNR: F/B $\uparrow$   \\
    \midrule
    PIFu    & $0.0792/0.0966$ & $0.8965/0.8742$ & $18.141/16.721$ & $0.0706/0.0849$ & $0.9242/0.9007$ & $20.104/17.926$ \\  
    GTA     & $0.0730/0.0891$ & $0.9003/0.8923$ & $18.790/18.229$ & $0.0633/0.0770$ & $0.9298/0.9275$ & $21.113/20.497$ \\
    ICON    & $0.0710/-$ & $0.8976/-$ & $18.613/-$ & $0.0608/-$ & $0.9291/-$ & $21.127/-$\\     
    ECON    & $0.0781/-$ & $0.8868/-$ & $18.454/-$ & $0.0658/-$ & $0.9261/-$ & $20.961/-$ \\ 
    SiFU    & $0.0692/0.0879$ & $0.9023/0.8915$ & $18.715/18.111$ & $0.0597/0.0768$ & $0.9302/0.9243$ & $21.101/20.349$ \\
    SiTH    & $0.0679/0.0843$ & $0.9007/0.8870$ & $18.417/17.608$ & $0.0612/0.0766$ & $0.9232/0.9107$ & $20.326/19.355$ \\

    \projecttitle    & $\textbf{0.0414/0.0643}$ & $\textbf{0.9603/0.9415}$ & $\textbf{22.347/20.849}$ & $\textbf{0.0457/0.0616}$ & $\textbf{0.9623/0.9512}$ & $\textbf{23.794/22.657}$ \\    
    \bottomrule
    \end{tabular*}
    }
    \vspace{-0.1cm}
    \caption{\textbf{Comparison of Human Texture with SOTA methods.} Note that only some methods predict the texture of the human body, so we render the textured 3D human reconstruction results of these methods in front view and back view, represented by ``F/B" symbols. The ICON and ECON methods only predict the front view texture. \label{main_exp_2d}}
    \vspace{-0.5cm}
\end{table*}

\myparagraph{Continuous remeshing} Building on inverse rendering research~\cite{botsch2004remeshing, palfinger2022continuous}, we iteratively refine the human mesh using the improved human normal maps ${\hat{I_{n}^{i}}}$. In each step, we differentiably render the predicted mesh's normal maps and compare them with ${\hat{I_{n}^{i}}}$ to calculate the loss and gradients. These gradients guide the adjustment of vertices and faces, allowing us to achieve a more detailed human mesh. The loss function is explicitly defined as:

\begin{equation}
    \mathcal{L}_{normal} = \sum_{i=0}^{N} \Vert \hat{I_{n}^{i}} - R(\mathcal{M'}, \textbf{k}) \Vert_{2}^{2}
\end{equation}
where the $\mathcal{L}_{normal}$ represents the main loss, and the $\mathcal{M}'$ is the iteratively predicted human mesh during the optimization, and $R(\cdot)$ function means the differentiable rendering.





\section{Experiments}
\label{sec:exp}

\subsection{Experiment Setup}
\myparagraph{Dataset} Previous studies often rely on datasets such as RenderPeople~\cite{renderpeople}, which limits reproducibility due to accessibility issues. For fair comparisons, similar to ICON~\cite{xiu2022icon}, we conduct experiments on the latest publicly available 3D human dataset, THuman2.0. Additionally, we use the CustomHumans~\cite{ho2023customhuman} and THuman3.0~\cite{thuman3.0} datasets for evaluation. During evaluation, all method utilizes the estimated SMPL/SMPL-X models as the body prior. Detailed information about datasets is in the supplementary material.


\subsection{Quantitative Comparison of Human Geometry}

\myparagraph{Evaluation Metrics} To evaluate the reconstruction quality, we follow the prior work SiTH~\cite{ho2024sith} to compute 3D metrics Chamfer distance (\textbf{CD}), Normal Consistency (\textbf{NC}), and \textbf{f-Score}~\cite{fscore} on our generated results. Moreover, we select the \textbf{LPIPS}~\cite{lpips}, \textbf{SSIM}, and \textbf{PSNR} as the 2D metrics to evaluate the texture quality of the reconstructed mesh. 

Table~\ref{main_exp_3d} highlights the superior performance of our method in reconstructing human geometry compared to other SOTA techniques. We evaluated all approaches on two OOD datasets, CustomHuman and THuman3.0, ensuring a fair comparison. Particularly, our method outperformed competitors, including the second-ranked method. On CustomHuman, we achieve improvements of $0.180/0.406$ on CD, $0.034$ on NC, and $6.277$ on f-score. For THuman3.0, the gains are $0.355/0.369$ on CD, $0.047$ on NC, and $9.861$ on f-score. We attribute our performance improvement to the introduction of our novel solution. Unlike traditional methods that focus solely on general human body geometry, our approach models human body geometry in a more detailed and multi-level manner. By enhancing human body reconstruction at three distinct levels, rather than treating it as a whole, we achieve superior results.


\subsection{Quantitative Comparison of Human Texture}


\begin{table*}[t!]
    \centering
    \renewcommand{\arraystretch}{0.85}
    \scalebox{0.86}{
    \begin{tabular*}{\textwidth}{@{\extracolsep{\fill}}lc|ccc|ccc}
    \toprule
    \multirow{2}{*}{Methods}& \multirow{2}{*}{Components}   & \multicolumn{3}{c}{CustomHuman} & \multicolumn{3}{c}{THuman3.0}  \\
                                                  &  ~   & \begin{tabular}{c} CD: P-to-S /\\ S-to-P $(\mathrm{cm}) \downarrow$ \end{tabular} & NC $\uparrow$ & f-score $\uparrow$ &
                                                                    \begin{tabular}{c} CD: P-to-S /\\ S-to-P $(\mathrm{cm}) \downarrow$ \end{tabular} & NC $\uparrow$ & f-score $\uparrow$  \\

    \midrule
    w/ 3-view Proj.    &  \multirow{5}{*}{SLE } & $1.634/1.790$ & 0.847 & 42.081 & $1.442/1.650$ & 0.829 & 45.525 \\ 
    w/ 2-view Proj.    &               & $1.772/ 2.061$ & 0.836 & 38.651 & $1.477/1.831$ & $0.820$ & $43.255$ \\
    w/ 1-view Proj.    &               & $1.842/2.313$ & 0.823 & 36.835 & $ 1.481/1.851$ & 0.819 & 42.802 \\  
    w/o Fouiour Proj. &               & $2.263/2.617$ & 0.806 & 31.624 & $1.966/2.160$ & 0.804 & 35.086 \\
    w/o Shape prior   &               & $2.130/2.595$ & 0.808 & 33.295 & $1.932/2.095$ & 0.798 & 37.685 \\
    \midrule 
    Ours ($\alpha$=0.25)  &  \multirow{3}{*}{JLA }  & $1.634/1.790$ & 0.847 & 42.081 & $1.442/1.650$ & 0.829 & 45.525 \\    
    Ours ($\alpha$=0.10)  &                                 & $1.650/1.809$ & 0.846 & 41.418 & $1.545/1.704$ & 0.824 & 44.008 \\     
    Ours ($\alpha$=0.00)  &                                 & $1.708/1.848$ & 0.843 & 40.429 & $1.611/1.746$ & 0.820 & 43.166 \\
    \midrule
    w/o WLR           &   WLR        & $1.634/1.790$ & 0.847 & 42.081 & $1.442/1.650$ & 0.829 & 45.525 \\  
    \midrule 
    Our Full Pipeline &      ~         & $\textbf{1.620}/\textbf{1.782}$ & \textbf{0.850} & \textbf{42.425} & $\textbf{1.408}/\textbf{1.633}$ & \textbf{0.834} & \textbf{46.091} \\  
    \bottomrule                     
    \end{tabular*}
    }
    \vspace{-0.1cm}
    \caption{\textbf{Ablation Study on Three Core Components.} We evaluate the effectiveness of the SLE module for reconstructing human geometry by comparing model performance with and without the 2D projection of 3D SMPL-X features. In the ``w/o Shape prior" setup, only the RGB image is used for reconstruction, serving as our baseline. In ``w/o Fourier Proj.," we encode and integrate 3D SMPL-X Fourier features directly with image features, bypassing the proposed Fourier space feature projection. For ``1-view Proj.," ``2-view Proj.," and ``3-view Proj.," we implement the proposed projection operation from one, two, and three camera views, respectively. We assess the JLA module's impact by comparing performance with and without human joint disturbance during training. The hyperparameter $\alpha$ controls augmentation intensity: ``$\alpha$= 0.00" indicates no disturbance, while ``$\alpha$ = 0.10" and ``$\alpha$ = 0.25" apply the strategy at varying intensities. Lastly, we evaluate the WLR module's effectiveness by testing conditions with and without this module.\label{abl_exp_3d}}
    \vspace{-0.2cm}
\end{table*}

In Table~\ref{main_exp_2d}, we also present a 2D quantitative comparison of human texture quality, highlighting our method's superiority over other SOTA approaches. On CustomHuman, our method enhances LPIPS by 0.0265/0.0200 (F/B), SSIM by 0.0580/0.0492 (F/B), and PSNR by 3.557/2.620 (F/B). Similarly, on THuman3.0, it improves LPIPS by 0.0140/0.015 (F/B), SSIM by 0.0321/0.0237 (F/B), and PSNR by 2.667/2.160 (F/B), achieving SOTA performance. We attribute the improvement in texture to our accurate reconstruction of human geometry. Once the geometric representation of the human body is enhanced, the overall quality naturally improves as well. This conclusion is demonstrated by the ablation study in Table~\ref{exp_abl}.

\begin{table}[!t]
\begin{center}
\scalebox{0.75}{
\begin{tabular}
{l|ccc}
\toprule

\multirow{2}{*}{Methods}  & \multicolumn{3}{c}{CustomHuman}  \\
&LPIPS: F/B $\downarrow$ & SSIM: F/B $\uparrow$ & PSNR: F/B $\uparrow$  \\

\midrule
    Ours (w/o SLE)       & $0.0497/0.0750$ & $0.9414/0.9205$ & $21.026/19.661$ \\
    Ours (w/o JLA)       & $0.0465/0.0666$ & $0.9496/0.9359$ & $21.535/20.327$ \\
    Our Full Pipeline   & $0.0414/0.0643$ & $0.9603/0.9415$ & $22.347/20.849$ \\

\midrule

& \multicolumn{3}{c}{THuman3.0} \\
&LPIPS: F/B $\downarrow$ & SSIM: F/B $\uparrow$ & PSNR: F/B $\uparrow$   \\

\midrule
     Ours (w/o SLE)   & $0.0575/0.0676$& $0.9408/0.9299$ & $22.129/21.048$ \\  
Ours (w/o JLA) & $0.0541/0.0657$ & $0.9449/0.9407$ & $22.262/21.647$ \\ 
Our Full Pipeline & $0.0457/0.0616$ & $0.9623/0.9512$ & $23.794/22.657$ \\    

\bottomrule
    \end{tabular}
}
\caption{\textbf{Ablation Study of SLE and JLA for Texture Quality.} We demonstrate through the ablation of SLE and JLA modules that the optimization of geometry, specifically the skeleton-level and joint-level, also leads to improvements in human texture. \label{exp_abl}}
    		\vspace{-0.8cm}

\end{center}
\end{table}

\begin{figure*}[t!]
    \centering
    \setlength{\belowcaptionskip}{-0.25cm}
    \includegraphics[width=0.98\linewidth]{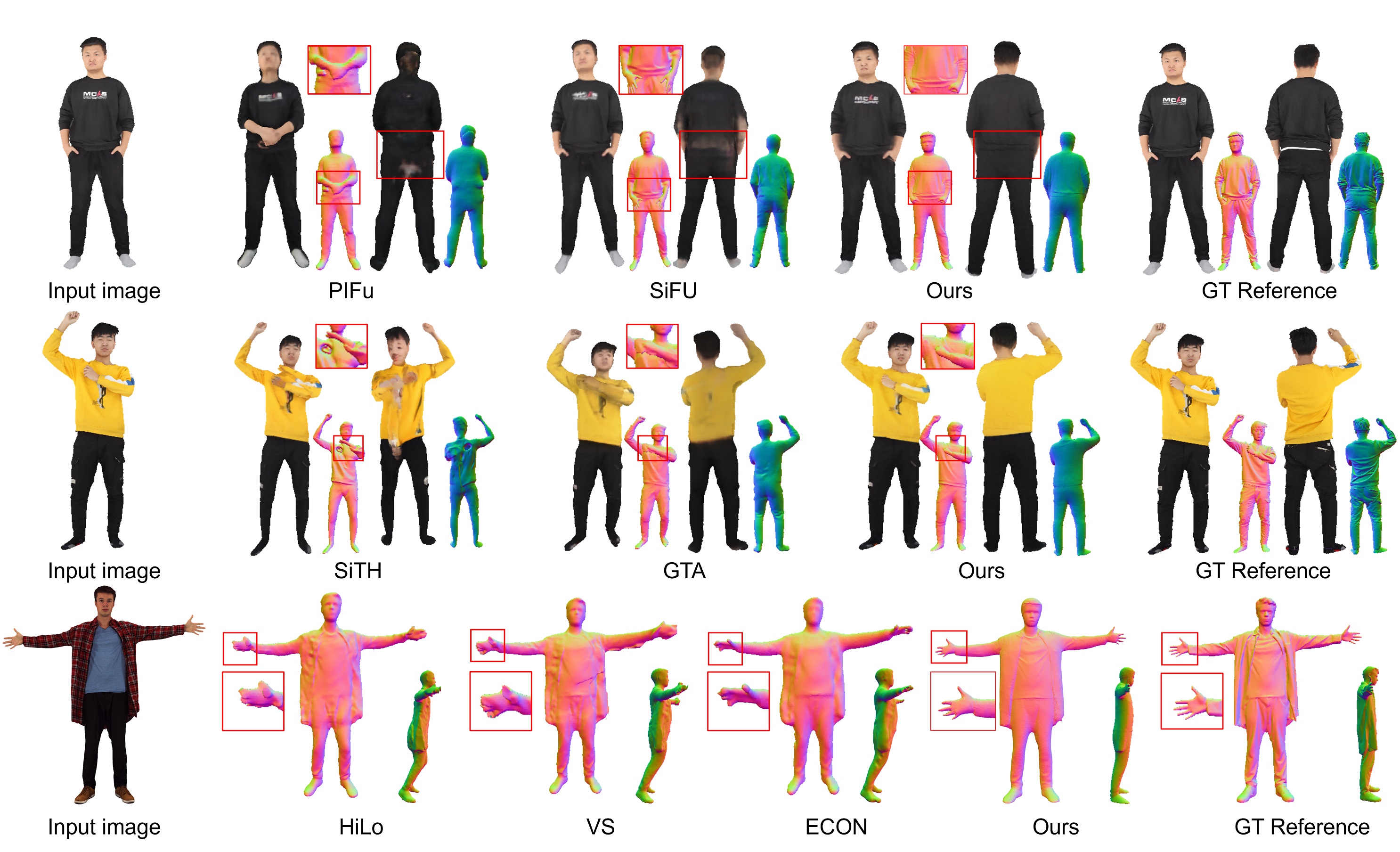}
    		\vspace{-0.4cm}
    \caption{\textbf{Qualitative comparison with SOTA methods.} In the first two rows, we present results from methods that generate both mesh and texture. The third row showcases methods that produce mesh only. Our proposed MultiGO significantly outperforms existing state-of-the-art methods in terms of skeleton integrity, posture accuracy, and wrinkling detail. In the first row, both PIFu and SiFU fail to represent human geometry and texture accurately, resulting in incorrect palm positioning and excess color at the back. In the second row, SiTH and GTA struggle with body geometry, particularly with overly slender arms or unnecessary protrusions. In contrast, our method distinctly highlights the contours of the human hand, outperforming other approaches. Please \textbf{zoom in} for more details.}
    \label{fig:vis}
    \vspace{-0.2cm}

\end{figure*}



\subsection{Quantitative Ablation Study}


\myparagraph{Effectiveness of SLE} Table~\ref{abl_exp_3d} highlights the significant impact of correctly integrating 3D features into the model. Unlike our proposed projection operation, directly encoding 3D features using methods like~\cite{3DShape2VecSet} and incorporating them into the reconstruction model yields poor results, even negatively affecting performance. This is likely due to the pre-trained model being optimized with extensive 2D RGB images, creating a modality and semantic gap between 2D and 3D features. The direct fusion of 3D SMPL-X and 2D image features is challenging with limited human data. Our method, however, enhances model performance by effectively leveraging geometric human priors. Increasing the number of mapping surfaces reveals further performance improvements, underscoring the method's effectiveness.

\myparagraph{Effectiveness of JLA} Table~\ref{abl_exp_3d} shows that JLA significantly enhances human geometry reconstruction. By fine-tuning disturbance intensity, model performance improves further. This suggests that training with augmented body meshes helps the model develop correction abilities, allowing accurate predictions despite slightly unsatisfactory estimated SMPL-X body mesh. This correction ability is rooted in effective image modality extraction and understanding.

\begin{figure}
    \centering
    \setlength{\belowcaptionskip}{-0.1cm}
    \includegraphics[width=1\linewidth]{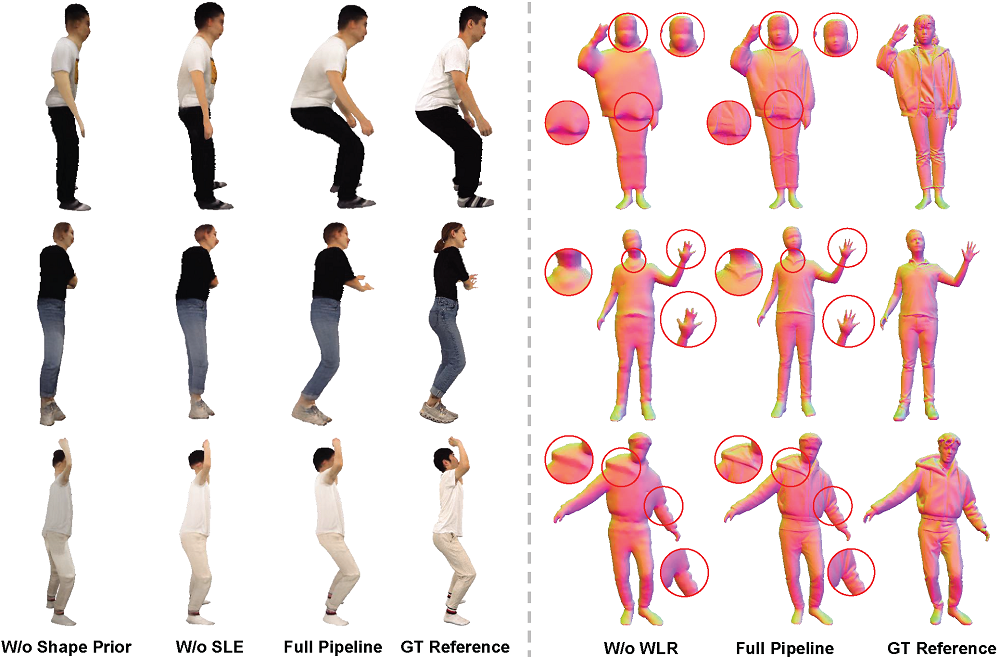}
    		\vspace{-0.3cm}

    \caption{\textbf{Visual Ablation.} \textit{Left}: The  SLE (``Full pipeline") makes the overall human skeleton close to the GT. \textit{Right}: The usage of WLR makes the wrinkles clearer. Please \textbf{zoom in} for more details.}
    \label{fig:abl_vis}
        		\vspace{-0.4cm}

\end{figure}

\myparagraph{Effectiveness of WLR} Table~\ref{abl_exp_3d} also highlights WLR's effectiveness in enhancing model performance across two test sets. Although the quantitative improvement is less pronounced than with SLE and JLA, WLR is crucial for refining details at the wrinkle level, such as face, and clothing, which are also essential for the task.

\myparagraph{Effect of SLE and JLA on texture quality} In Table~\ref{exp_abl}, we observe that the proposed SLE and JLA models, aimed at enhancing the geometric quality of the human body at both the skeleton and joint levels, also lead to improvements in body texture quality. This correlation is intuitive: when our reconstructed human body is geometrically aligned with the target, the rendered 2D image aligns more closely as well, resulting in better 2D performance.
\subsection{Visualiztion}
\label{sec:label}

\myparagraph{Comparison with SOTA methods} In Figures~\ref{fig:enter-label} and~\ref{fig:vis}, we intuitively demonstrate the superiority of our algorithm over existing methods. The figure clearly shows that our approach excels in capturing human skeletons, joints, and wrinkles, thanks to our multi-level modeling of human geometry.   Additionally, this geometric improvement enables our method to achieve SOTA performance in texture as well.

\myparagraph{Effect of SLE and JLA on reconstruction quality} Figure~\ref{fig:abl_vis} illustrates the effectiveness of our proposed modules. On the left subfigure, it is found that without any geometric prior knowledge, the generated human geometry and texture are inadequate. Although integrating 3D SMPL-X information improves the geometric quality, the enhancement is minimal. In contrast, our SLE module can effectively align the generated human body with the target quality. On the right subfigure, we observe that the generated mesh lacks fine details like wrinkles without our proposed WLR module. However, applying the WLR largely enhances the detail and realism of the human mesh.

\section{Conclusion}
\label{sec:conclusion}
In this paper, we introduce MultiGO, a novel framework for monocular 3D human reconstruction that overcomes limitations in handling human-specific geometry. By incorporating a Skeleton-Level Enhancement module, Joint-Level Augmentation strategy, and Wrinkle-Level Refinement module, MultiGO leverages body geometric priors to enhance reconstruction quality across various granularity levels. This approach optimizes large object Gaussian models to address human geometry challenges, resulting in more accurate and detailed reconstructions. Extensive experiments on two test sets demonstrate that our method achieves SOTA performance.
{
    \small
    \bibliographystyle{ieeenat_fullname}
    \bibliography{main}
}


\end{document}